%% file: arxiv.tex
\pdfoutput=1 

\documentclass[sigconf]{acmart}

\usepackage[hyphenbreaks]{breakurl}

\usepackage[ruled,vlined]{algorithm2e}
\usepackage{pgfplots}

\pgfplotsset{compat=newest}
\usepgfplotslibrary{groupplots}
\usepgfplotslibrary{dateplot}

\usepackage{bbm}
\usepackage{tikz}

\usepackage{soul}
\usepackage{amsfonts, amsmath, amsthm, graphicx, verbatim}
\usepackage{xcolor}

\usepackage{multirow}
\usepackage{subcaption}

\usepackage{tabularx}
\usepackage{booktabs}
\usepackage{mathtools}
\usepackage{multirow}

\usepackage{standalone}

\newtheorem{corollary}{Corollary}
\newtheorem{proposition}{Proposition}
\newtheorem{definition}{Definition}

\AtBeginDocument{%
  \providecommand\BibTeX{{%
    \normalfont B\kern-0.5em{\scshape i\kern-0.25em b}\kern-0.8em\TeX}}}

\copyrightyear{2021}
\acmYear{2021}
\setcopyright{acmlicensed}\acmConference[COMPASS '21]{ACM SIGCAS Conference on Computing and Sustainable Societies (COMPASS)}{June 28--July 2, 2021}{Virtual Event, Australia}
\acmBooktitle{ACM SIGCAS Conference on Computing and Sustainable Societies (COMPASS) (COMPASS '21), June 28--July 2, 2021, Virtual Event, Australia}
\acmPrice{15.00}
\acmDOI{10.1145/3460112.3471950}
\acmISBN{978-1-4503-8453-7/21/06}

\begin{document}\sloppy
\pdfoutput=1

\title{A Ranking Approach to Fair Classification}

\author{Jakob Schoeffer}
\affiliation{%
  \institution{Karlsruhe Institute of Technology}
  \city{Karlsruhe}
  \country{Germany}
}
\email{jakob.schoeffer@kit.edu}

\author{Niklas Kuehl}
\affiliation{%
  \institution{Karlsruhe Institute of Technology}
  \city{Karlsruhe}
  \country{Germany}
}
\email{niklas.kuehl@kit.edu}

\author{Isabel Valera}
\affiliation{%
  \institution{Saarland University}
  \city{Saarbrücken}
  \country{Germany}
}
\email{ivalera@cs.uni-saarland.de}


\begin{abstract}
    Algorithmic decision systems are increasingly used in areas such as hiring, school admission, or loan approval. 
    Typically, these systems rely on labeled data for training a classification model.
    However, in many scenarios, ground-truth labels are unavailable, and instead we have only access to \emph{imperfect} labels as the result of (potentially biased) human-made decisions. 
    Despite being imperfect, historical decisions often contain \emph{some} useful information on the unobserved true labels. 
    In this paper, we focus on scenarios where only imperfect labels are available and propose a new fair ranking-based decision system based on monotonic relationships between legitimate features and the outcome.
    %
    Our approach is both intuitive and easy to implement, and thus particularly suitable for adoption in real-world settings.
    %
    More in detail, we introduce a distance-based decision criterion, which incorporates useful information from historical decisions and accounts for unwanted correlation between protected and legitimate features.
    Through extensive experiments on synthetic and real-world data, we show that our method is fair in the sense that a) it assigns the desirable outcome to the most qualified individuals, and b) it removes the effect of stereotypes in decision-making, thereby outperforming traditional classification algorithms.
    %
    Additionally, we are able to show theoretically that our method is consistent with a prominent concept of individual fairness which states that ``similar individuals should be treated similarly.''
    %
\end{abstract}

\ccsdesc[500]{Computing methodologies~Machine learning}
\ccsdesc[500]{Computing methodologies~Philosophical/theoretical foundations of artificial intelligence}
\ccsdesc[500]{Social and professional topics~Socio-technical systems}
\ccsdesc[500]{Information systems~Decision support systems}

\keywords{Fairness, bias, classification, ranking, algorithmic decision-making} 

\maketitle


\section{Introduction}\label{sec:introduction}
Algorithmic decision systems have been increasingly used for decision support in recent years.
A common perception is that algorithms can avoid human bias and make more objective and transparent decisions \cite{eustudy2019}.
However, as algorithms support humans with evermore consequential decisions, they have also become subject to enhanced scrutiny.
In 2016, journalists at ProPublica found that COMPAS, a system used by US courts to assess defendants' risk of recidivism, was unfair towards black people \cite{angwin2016machine}.
In November 2019, Bloomberg reported that Steve Wozniak was suspecting the algorithm that determines credit limits for Apple's credit card to discriminate against women \cite{nasiripour2019}.
These and other examples make obvious the need for understanding root causes and developing techniques to combat algorithmic unfairness.
In large part, prior work has focused on formalizing the concept of fairness and enforcing certain statistical equity constraints when making predictions---mostly in a setting of binary classification, for instance when it must be decided whether a loan should be offered or not.

However, traditional classification algorithms require access to \emph{actual} ground-truth labels, which are often unavailable \cite{lakkaraju2017selective}.
In practice, we may only have access to \emph{imperfect} labels, generally as the result of (potentially biased) historical human-made decisions.
Inspired by the argumentation of \citet{kilbertus2020fair}, we propose to \emph{not} learn to predict imperfect labels.
Instead, we introduce a meritocratically fair decision criterion based on an observation's distance to what we call the \emph{North Star}---a (potentially hypothetical) observation that is most qualified in a given scenario.
Our approach induces both a ranking and an opportunity to classify observations, based on monotonic relationships between legitimate features and the outcome.
We also put forward ideas to a) incorporate useful information from historical decisions (Section \ref{sec:extracting_useful_information}) and b) reduce the importance of features that are highly correlated with protected attributes (e.g., gender) in the decision-making process (Section \ref{sec:accounting_for_relationships}).

The rest of the paper is structured as follows: In Section \ref{sec:background}, we introduce important concepts and related work.
Section \ref{sec:proposed_methodology} represents the core of our work---the methodology as well as theoretical results.
In Section \ref{sec:case_study}, we illustrate our method by the example of the German Credit data set, and we conduct extensive experiments on synthetic data in Section \ref{sec:experiments_synthetic}.
Section \ref{sec:conclusion} summarizes our work, discusses its limitations, and provides potential areas for follow-up work.
All code and data is openly available through GitHub.

\section{Background}\label{sec:background}
To lay out the foundations, we briefly introduce important concepts related to our proposed methodology.
We start with a summary of the notation used in this paper.
We call $A$ the set of \emph{protected} features which must not be discriminated against, and let $a_k \in A,\ k \in \{1,\dots,K\}$ be the individual protected features.
The US Equality Act \cite{usequalityact}, for instance, defines sex, gender identity, and sexual orientation, among others, as protected features.
In line with other related work, we assume that the decision whether a feature is protected or not is made externally \cite{zliobaite2015survey}.
We further define $x_\ell \in X,\ \ell \in \{1,\dots,L\}$ as the non-protected (or \emph{legitimate}) features, and $Y$ as the set of \emph{imperfect} labels, which can be either positive ($+$) or negative ($-$).
We call these labels \emph{imperfect} as they are only a noisy signal of the true labels.
A given data set consisting of $A$, $X$, $Y$ is referred to as $\mathcal{D}$.
Lastly, we call $\widehat{Y}$ the predictor, a function that maps observations to positive or negative outcomes.
When referring to an individual observation, we use superscripts like $A^{(i)}$, which would be the set of protected features for observation $(i)$, while we have $N$ observations in total.

\subsection{Relevant notions of fairness}\label{sec:notions_of_fairness}
\citet{mehrabi2019survey} define fairness in the context of decision-making as the ``absence of any prejudice or favoritism towards an individual or a group based on their intrinsic or acquired traits.''
Generally, existing literature distinguishes individual from group fairness definitions.
In this work, we are primarily concerned with individual fairness---however we also make group fairness-related arguments in the spirit of demographic parity \cite{zafar2017fairness} later on.
%
%
%
A typical approach aiming at individual fairness is \emph{fairness through awareness} (FTA) \cite{dwork2012fairness}. We will briefly introduce this, as well as the conception of \emph{fairness through unawareness} (FTU) \cite{grgic2016case}, due to their importance for our work.

In line with \citet{grgic2016case}, FTU is the conception that an algorithm is fair so long as any protected features are not explicitly used in the decision-making process.
%
In other words, FTU simply requires a predictor $\widehat{Y}$ to ignore all protected features in $A$.
However, as \citet{hardt2016equality} argue, this definition is ineffective in the presence of strong correlation between protected and legitimate features.
In the work at hand, we address this issue by penalizing highly correlated features with respect to their importance for decision-making.

According to \citet{dwork2012fairness}, FTA says that an algorithm is fair if it gives similar predictions to similar individuals.
%
Formally, FTA requires an appropriate distance metric $d(\cdot,\cdot)$.
If, for two individuals $(i)$ and $(j)$, $d(i,j)$ is small, then FTA requires that $\widehat{Y}^{(i)}\approx \widehat{Y}^{(j)}$.
As stated by \citet{dwork2012fairness}, the main challenge with this notion is defining an appropriate distance metric.
In most cases, this requires domain-specific knowledge.

\subsection{Related work}
Most of existing work on algorithmic fairness has been concerned with fair classification.
Herein, numerous articles have been published on how to formally define \cite{grgic2016case,hardt2016equality,dwork2012fairness,kusner2017counterfactual,corbett2017algorithmic,pedreshi2008discrimination} and enforce \cite{kilbertus2020fair,zafar2017fairness,zafar2017fairnessconstraints,hardt2016equality,kamiran2009classifying,kamiran2012data,kamishima2012fairness,calmon2017optimized} fairness.
Generally, fairness-aware techniques can be divided into three categories which are based on the process step of the application: The first category is concerned with removing existing bias from the training data (pre-processing).
Typical approaches involve transformation of the data \cite{calmon2017optimized} or changing the class labels for training \cite{kamiran2012data}.
The second category involves modification of existing algorithms (in-processing), typically through adding fairness constraints \cite{zafar2017fairnessconstraints} or penalizing discrimination, for instance by means of regularization \cite{kamishima2012fairness}.
The third category includes all techniques aimed at changing the output of a potentially unfair model (post-processing).
\citet{hardt2016equality}, for instance, construct a non-discriminating predictor from an existing one via solving an optimization problem.
However, if ground-truth labels are not (or selectively) available \cite{lakkaraju2017selective}, then maximizing for prediction accuracy seems counterintuitive and is, in fact, suboptimal \cite{kilbertus2020fair}.

More recently, alternative concepts based on the theory of causal inference have evolved \cite{kilbertus2017avoiding,kusner2017counterfactual}.
While these approaches have shown promising results, they generally make strong assumptions about the causal structure of the world.
An in-depth discussion (including common misconceptions) of causal models in the realm of algorithmic fairness is provided, for instance, by \citet{hu2020s}.

Fair ranking approaches can be split into pre-processing \cite{yang2017measuring}, in-processing \cite{zehlike2020reducing}, and post-processing \cite{biega2018equity,celis2018ranking,singh2017equality,singh2018fairness,zehlike2017fa} techniques as well \cite{castillo2019fairness}.
With respect to quantifying fairness, most of existing methods apply an attention-based criterion, aiming at equalizing exposure of observations in, for instance, (web) searches \cite{biega2018equity,singh2017equality,singh2018fairness}.
Furthermore, a majority of existing literature has been focusing on achieving group fairness, whereas individual fairness considerations for rankings remain scarce---with few exceptions, such as work by \citet{biega2018equity}, where the authors introduce a mechanism to achieve individual fairness across a \emph{series} of rankings.
To the best of our knowledge, no existing work on fair ranking is closely related to ours in terms of methodology.

Perhaps the most related work is an article by \citet{wang2020deontological}.
Here, the authors put forward the idea of optimizing classification accuracy subject to a classifier being monotonic in a given set of features.
Thereby, it is argued, the classifier can evade violating ``common deontological ethical principles and social norms such as [\dots] `do not penalize good attributes'.''
While the idea of enforcing monotonicity constraints is similar, we have identified three major differences to our work:
\begin{enumerate}
    \item \citet{wang2020deontological} use supervised machine learning to predict ground-truth labels, whereas we assume imperfect labels.
    Our proposal in the case of imperfect labels is to \emph{not} maximize for accuracy in the first place.
    \item They do not take measures to prevent the algorithm from ``exploiting'' protected information to achieve higher accuracy.
    \item They do not account for the well-known problem of indirect discrimination, which occurs when (seemingly) legitimate features are highly correlated with protected features.
\end{enumerate}

\section{Proposed methodology}\label{sec:proposed_methodology}
In this chapter, we introduce our proposed ranking algorithm for decision-making with imperfectly labeled data, that is, the common case where ground-truth labels are not available.
Specifically, we assume we are given data $\mathcal{D}$ with imperfect labels stemming from human-made decisions, for instance whether an applicant was admitted to graduate school or not.

\begin{table}
  \caption{Exemplary graduate school admission data.}
  \label{tab:grad_school_data_set}
  \centering
  \begin{tabular}{cccccc}
    \toprule
    ID & Gender      & GRE V     & GRE Q    & GRE AW    & $Y$ \\
    \midrule
    1 & \emph{male} & 147 & 144 & 3.0 & $(+)$ \\
    2 & \emph{male} & 146 & 140 & 3.5 & $(+)$ \\
    \dots & \dots & \dots & \dots & \dots & \dots \\
    11 & \emph{female} & 153 & 147 & 3.5 & $(-)$ \\
    \dots & \dots & \dots & \dots & \dots & \dots \\
    \bottomrule
  \end{tabular}
\end{table}

Our approach follows a notion of individual fairness that aims at uniting both fairness definitions from Section \ref{sec:notions_of_fairness}, FTU and FTA. Note that this idea is closely related to the concept of ``meritocratic fairness,'' as coined by \citet{kearns2017meritocratic}.
We call an algorithm \emph{meritocratically fair} if it assigns the positive outcome to the most qualified observations, regardless of protected features.
%
%
This definition is in line with many equal employment opportunity policies, yet disregarding affirmative action.
Based on this notion, we can also define meritocratic \emph{un}fairness:

\begin{definition}[Meritocratic unfairness]\label{def:indiv_unfairness}
An individual observation $(i)$ is treated unfairly over a different observation $(j)$ if $(i)$ is more qualified than $(j)$ but: a) ranked lower, or b) assigned $(-)$ while $(j)$ is assigned $(+)$.
\end{definition}

We will use Definition \ref{def:indiv_unfairness} for evaluation purposes later on.
However, a definition of what \emph{qualified} means can hardly be given without knowledge of the respective use case---a viewpoint that is shared, among others, by \citet{dwork2012fairness}.
We will address this now.

To illustrate our ideas, we construct a (simplified) synthetic graduate school admission data set and use it as a running example---an excerpt is shown in  Table \ref{tab:grad_school_data_set}.
The data set consists of 1,000 observations, with 50\% being males and 50\% females (protected feature).
The GRE scores (legitimate features) are three-dimensional: GRE V(erbal Reasoning), GRE Q(uantitative Reasoning), GRE A(nalytical) W(riting). We sample them from multivariate Gaussian distributions $\mathcal{N}(\mu_{m(ale)},\Sigma_m)$ and $\mathcal{N}(\mu_{f(emale)},\Sigma_f)$, where
\begin{equation*}
    \mu_m=[150.7,156.1,3.5],\quad \mu_f=[150.3,151.2,3.7],
\end{equation*}
and the covariance matrices
\begin{equation*}
    \Sigma_m=
    \begin{bmatrix}
    81.00 & 28.15 & 5.43 \\
    28.15 & 84.64 & 1.16 \\
    5.43 & 1.16 & 0.81
    \end{bmatrix},\ 
    \Sigma_f=
    \begin{bmatrix}
    65.61 & 24.51 & 4.34 \\
    24.51 & 79.21 & 1.00 \\
    4.34 & 1.00 & 0.64
    \end{bmatrix}
\end{equation*}
are derived from the official data provided by the administrator of the GRE test \cite{ets19,ets18}.
For compliance with the official ranges of scores, we round and truncate the sampled scores such that a) GRE V and GRE Q scores are between 130 and 170 in one-point increments, and b) GRE AW scores are between 0 and 6 in half-point increments.
To simulate historical admission decisions, we scale the legitimate features between 0 and 1, and generate imperfect labels as
\begin{equation*}\label{eq:synthetic_labels_1}
    Y =
    \begin{cases}
        (+) & \textbf{if } 0.1\cdot \mathbbm{1}_{male} + 0.2\cdot \text{GRE V} + 0.5\cdot \text{GRE Q} \\
        &\quad + 0.2\cdot \text{GRE AW} + \epsilon > 0.5 \\
        (-) & \textbf{otherwise},
    \end{cases}
\end{equation*}
where $\mathbbm{1}$ is the indicator function and $\epsilon \sim \mathcal{U}(0,0.1)$ is noise.
Note that male applicants are given an unfair advantage over their female counterparts.
Apart from this, the high importance of GRE Q scores could be representative of a technical university's admission process.

We assume that for any specific use case, we are given (e.g., by an expert) or can easily derive the information of how any legitimate and relevant feature should impact the final decision---specifically, whether higher ($\uparrow$) or lower ($\downarrow$) values are beneficial with respect to the positive outcome.
Note that if certain feature interactions have a known monotonic relationship with the outcome, then these interactions can be added as additional features and assigned a ($\uparrow$) or ($\downarrow$) as well.
Certainly, in many cases these dependencies are obvious and need not be verified by an expert.
For instance, in our graduate school example, it is clear that high GRE scores are more beneficial towards being admitted than low scores.
Alternatively, if obtaining this information from an expert is too expensive, we could potentially infer the ($\uparrow$) or ($\downarrow$) relationships from $\mathcal{D}$.
The idea that relevant features should have a monotonic relationship with the outcome is, for instance, similarly introduced by \citet{wang2020deontological}.

With this information, we first scale the legitimate features $X$ such that all values are in $[0,1]$.
We call the scaled legitimate features $z_\ell \in Z,\ \ell \in \{1,\dots,L\}$.
We further require that the probability of the positive outcome increases with the value of any $z_\ell$.
For that, we perform $z_\ell \leftarrow (1-z_\ell)$ if the original relationship between $x_\ell$ and the outcome is ($\downarrow$).
These steps are summarized in Algorithm \ref{algo:scaling}.
Note that we can also apply Algorithm \ref{algo:scaling} to observations that are not contained in $\mathcal{D}$---in that case, we need to assume that the resulting values of $z$ are capped at 0 and 1.

\begin{algorithm}[t]
\SetKwInOut{Input}{Input}
\SetKwInOut{Output}{Output}
\SetAlgoLined
\Input{Legitimate features $x_1,\dots,x_L$ of $\mathcal{D}$, including ($\uparrow$) or ($\downarrow$) relationships.}
\Output{Scaled features $z_1,\dots,z_L$.}
\For{$\ell \in \{1,\dots,L\}$}{
    $z_\ell^{(i)} \leftarrow \frac{\left(x_\ell^{(i)} - \min_{j} x_\ell^{(j)}\right)}{\left(\max_{j} x_\ell^{(j)} - \min_{j} x_\ell^{(j)}\right)}$ for all $i \in \{1,\dots,N\}$\;
    \If{$x_\ell$ is ($\downarrow$)}{
        $z_\ell \leftarrow (1-z_\ell)$\;}
}
\caption{Scaling of legitimate features.}
\label{algo:scaling}
\end{algorithm}

\subsection{Measuring distance to the North Star}\label{sec:measuring_distance}
Our idea is to fairly rank observations based on their distance to what we call the \emph{North Star}.

\begin{definition}[North Star]
Given a data set $\mathcal{D}$ and the respective legitimate features $z_\ell,\ \ell \in \{1,\dots,L\}$ scaled as in Algorithm \ref{algo:scaling}, the \emph{North Star} is a (potentially hypothetical) observation $(\star)$ that attains the maximum observed value for each legitimate feature:
\begin{equation}
    z_\ell^{(\star)}:=\max_{i \in \{1,\dots,N\}} z_\ell^{(i)} = 1\quad \forall \ell \in \{1,\dots,L\}.
\end{equation}
\end{definition}
Now, we can compute the distance of single observations to the North Star.
For that, we choose the taxicab metric for its clear interpretation and its favorable behavior in higher dimensions \cite{aggarwal2001surprising}.
Note that the approach also works for other metrics.
We define the distance of an observation $(i)$ to the North Star as follows:
\begin{equation}\label{eq:distance}
    d(i,\star) := \sum_{\ell=1}^L \left(1-z_\ell^{(i)}\right),
\end{equation}
considering that $z_\ell^{(i)}\in [0,1]$ for all $\ell$ and observations $(i)$.
Note that we assume symmetry of our distance measure, that is, $d(i,\star)=d(\star,i)$.
In our example, this distance would be 0 for applicants with the perfect scores of $\text{GRE V} = 170$, $\text{GRE Q} = 170$, and $\text{GRE AW} = 6.0$.

\subsection{Extracting useful information from historical decisions}\label{sec:extracting_useful_information}
In a next step, we enhance the distance formula in Equation (\ref{eq:distance}) with useful information from historical data.
Despite the fact that our given data $\mathcal{D}$ contains only imperfect labels, we argue that historical decisions often contain \emph{some} useful information on the true labels.
Specifically, we aim to extract the relative importance of legitimate features from historical decisions, assuming that important features from the past are still important at present.
In our running example, for instance, we know that labels are biased, but we still want to capture that GRE Q scores are most important for admission at a technical university.

Our rationale is the following: While Equation (\ref{eq:distance}) implicitly treats every feature as having equal importance, we want to account for the fact that some features are undoubtedly more important than others for decision-making.
Even though we could explicitly ask experts for this information, similar to the ($\uparrow$) or ($\downarrow$) relationships, we argue that manually quantifying the importance of individual features (in \%) is often intractable.
We, therefore, propose to learn these importances directly from $\mathcal{D}$, for instance through the concept of permutation importance \cite{breiman2001random}---which is defined as the decrease in model score when the value of this respective feature is randomly permuted.
The result is an estimate of how much a given model depends on this feature.
For that, we train a classifier on $\mathcal{D}$ and obtain feature importances $\omega_1,\dots,\omega_L$, with $\omega_1,\dots,\omega_L\geq 0$ and $\sum_{\ell=1}^L \omega_\ell = 1$ for all legitimate features.\footnote{It might happen that the legitimate features cannot predict the historical labels reasonably well (e.g., if labels are random). In such cases we can skip this step.}
In a next step, we can now adjust Equation (\ref{eq:distance}) by adding $\omega_1,\dots,\omega_L$ as weights to reflect the importance of each legitimate feature:
\begin{equation}\label{eq:distance_mod_1}
    d'(i,\star) := \sum_{\ell=1}^L \omega_\ell\left(1-z_\ell^{(i)}\right).
\end{equation}
In the running example, we obtain $\omega_{V}=0.10$, $\omega_{Q}=0.73$, and $\omega_{AW}=0.17$, with standard deviations $\sigma_V=0.006$, $\sigma_Q=0.012$, and $\sigma_{AW}=0.005$, by fitting a random forest classifier with bootstrapping and using the \texttt{permutation\_importance} function of \texttt{scikit-learn} \cite{scikit-learn}.
This means that $d'$ would be most sensitive to changes in GRE Q scores, as desired.

\subsection{Accounting for relationships between legitimate and protected features}\label{sec:accounting_for_relationships}
As outlined in Section \ref{sec:notions_of_fairness}, as well as by \citet{hardt2016equality} and \citet{pedreshi2008discrimination}, the fundamental weakness of FTU as a notion of fairness is the fact that protected features can sometimes be predicted from legitimate features.
It is particularly problematic if legitimate features are highly correlated with protected features---we account for this by penalizing high correlation.

To measure general monotonic (i.e., not just linear, as for instance Pearson's $r$ would do) relationships between two data samples, we can use Spearman's rank correlation coefficient (SRCC).
For the rankings $rk_a$ and $rk_x$ of two samples $a$ and $x$, SRCC $\rho_{a,x}$ is calculated as follows:
\begin{equation}\label{eq:rho}
    \rho_{a,x} = \frac{\textrm{cov}(rk_a,rk_x)}{\sigma_{rk_a}\sigma_{rk_x}},
\end{equation}
where cov is the covariance and $\sigma$ the standard deviation.
An SRCC of $\pm1$ occurs if one sample is a perfect monotonic function of the other.
Using Equation (\ref{eq:rho}), we can then compute:
\begin{equation}\label{eq:max_rho}
    \widetilde{\rho}_\ell := \max_{k \in \{1,\dots,K\}} \left\{\left|\rho_{a_k,z_\ell}\right|\right\}\quad \forall \ell \in \{1,\dots,L\},
\end{equation}
as the maximum absolute rank correlation between a given legitimate feature $z_\ell$ and any protected feature $a_k$.
We take the absolute values of SRCC in order to have $\widetilde{\rho}_\ell \in [0,1]$.
However, our idea would also be consistent with, for instance, squaring the SRCC values instead.
Our intuition behind taking the maximum over, for instance, the sum is that we do \emph{not} want to penalize having many low individual absolute correlations---but rather scenarios where a seemingly legitimate feature is a (potentially noisy) proxy for one of the protected features.

Note that SRCC works for both numerical and ordinal features---this is important for our work.
%
If, however, non-binary categorical features are present in $\mathcal{D}$, then traditional correlation measures are generally not a meaningful way of determining relationships.
Alternatively, for instance, we might want to refer to the correlation ratio \cite{pearson1911correction}, usually denoted by $\eta \in [0,1]$, which measures the relationship between \emph{inter-}category variability and \emph{intra-}category variability of some feature.
As an example, assume we have a three-dimensional feature $\textit{gender} \in \{F,M,O\}$ and want to quantify the relationship between gender and GRE V as well as GRE Q.
Further assume that we have three observations per category (i.e., gender) and that the feature values are given as in Table \ref{tab:correlation_ratio}.
By construction, the overall variability of GRE V scores is solely due to inter-category variability ($\eta=1$), whereas for GRE Q the category means are the same, hence $\eta=0$.
\begin{table}
  \caption{Exemplary data for illustrating correlation ratio.}
  \label{tab:correlation_ratio}
  \centering
  \begin{tabular}{l c c c }
    \toprule
    ID & Gender & GRE V & GRE Q \\
    \midrule
    1 & \emph{F} & 130 & 140 \\
    2 & \emph{F} & 130 & 150 \\
    3 & \emph{F} & 130 & 160 \\
    4 & \emph{M} & 150 & 140 \\
    5 & \emph{M} & 150 & 150 \\
    6 & \emph{M} & 150 & 160 \\
    7 & \emph{O} & 170 & 140 \\
    8 & \emph{O} & 170 & 150 \\
    9 & \emph{O} & 170 & 160 \\
    \bottomrule
  \end{tabular}
\end{table}
Because of the same value range and corresponding interpretation of $\eta$ and $\left|\rho_{a_k,z_\ell}\right|$, we could straightforwardly adapt Equation (\ref{eq:max_rho}) by replacing the latter with the former.

For simplicity, and because most traditional classification algorithms require encoding of categorical features as well, we assume in the following that $\mathcal{D}$ does \emph{not} contain non-binary categorical features (e.g., because any such feature has been encoded accordingly) and that SRCC is applicable.
We can then use $\widetilde{\rho}_\ell$ to further adjust the distance measure from Equation (\ref{eq:distance_mod_1}):
\begin{equation}\label{eq:distance_mod_2}
    d''(i,\star) := \sum_{\ell=1}^L \omega_\ell \left(1-\widetilde{\rho}_\ell\right) \left(1-z_\ell^{(i)}\right),
\end{equation}
where high values of $\widetilde{\rho}_\ell$ reduce the importance of $z_\ell$ on the distance $d''$.
Note that in the extreme case of $\widetilde{\rho}_\ell=1$, the distance $d''$ will be independent of feature $z_\ell$.
This is desirable as it renders ineffective the possibility of introducing proxies for protected features under seemingly innocuous names.

For our graduate school admission example, we calculate the SRCC values using the \texttt{spearmanr} function of \texttt{SciPy} \cite{2020SciPy-NMeth}.
Note that we only have one protected feature---\emph{gender}.
The absolute correlations are $\widetilde{\rho}_V=0.035$, $\widetilde{\rho}_Q=0.262$, and $\widetilde{\rho}_{AW}=0.167$.
While these values are not strikingly high, we may infer that there is a stronger relationship between \emph{gender} and GRE Q than with GRE V or GRE AW.
The importance of GRE Q for admission is thus reduced by 26.2\%, as opposed to 3.5\% and 16.7\% for GRE V and GRE AW, respectively.
In general, even if a seemingly legitimate feature was highly important for past decisions, its importance will vanish if it is highly correlated with a protected feature, as desired.

Coming back to Definition \ref{def:indiv_unfairness}, we now define what \emph{being more qualified} could mean:

\begin{definition}[Higher qualification]\label{def:qualification}
We call an observation $(i)$ \emph{more qualified than} $(j)$ if, according to the $(\uparrow)$ or $(\downarrow)$ relationships between features and positive outcome, $(i)$ is better or equal than $(j)$ for all legitimate features and strictly better for at least one $\ell' \in \{1,\dots,L\}$, with $\omega_{\ell'}\neq 0$ and $\widetilde{\rho}_{\ell'}\neq 1$.
\end{definition}

Note that \emph{being more qualified} is a stronger requirement than observation $(i)$ having a shorter distance to the North Star than $(j)$, that is, \emph{being more qualified} implies a shorter distance to the North Star.
The converse is not generally true.
This implication is formally stated in the following proposition:

\begin{proposition}\label{theorem:more_qualified_shorter_distance}
If, according to Definition \ref{def:qualification}, an observation $(i)$ is more qualified than observation $(j)$, then $d''(i,\star)$ is strictly smaller than $d''(j,\star)$, where $d''$ is defined as in Equation (\ref{eq:distance_mod_2}).
\end{proposition}

\begin{proof}
Assume $(i)$ is more qualified than $(j)$, and w.l.o.g. assume that all legitimate features are scaled as in Algorithm \ref{algo:scaling}. Then we have:
\begin{equation*}
    z_\ell^{(i)} \geq z_\ell^{(j)}\ \forall \ell \in \{1,\dots,L\}\quad \text{and}\quad \exists \ell'\in \{1,\dots,L\}: z_{\ell'}^{(i)} > z_{\ell'}^{(j)}.
\end{equation*}
With $\psi_\ell:=\omega_\ell \left(1-\widetilde{\rho}_\ell\right)\in [0,1]$ and $\psi_{\ell'}\neq 0$, we then obtain:
\begin{align*}
    d''(i,\star)&=\sum_{\ell=1}^L \psi_\ell \left(1-z_\ell^{(i)}\right)\\
    &=\sum_{\ell=1}^L \psi_\ell - \left(\psi_1 z_1^{(i)}+\dots+\psi_{\ell'} z_{\ell'}^{(i)}+\dots+\psi_L z_L^{(i)}\right)\\
    &< \sum_{\ell=1}^L \psi_\ell - \left(\psi_1 z_1^{(j)}+\dots+\psi_{\ell'} z_{\ell'}^{(j)}+\dots+\psi_L z_L^{(j)}\right)\\
    &=d''(j,\star),
\end{align*}
since $\psi_{\ell'} z_{\ell'}^{(i)}>\psi_{\ell'} z_{\ell'}^{(j)}$ and $\psi_{\ell} z_{\ell}^{(i)}\geq \psi_{\ell} z_{\ell}^{(j)}$ for all other $\ell \in \{1,\dots,L\}\setminus \{\ell'\}$.
\end{proof}

Note that in Table \ref{tab:grad_school_data_set}, according to Definitions \ref{def:indiv_unfairness} and \ref{def:qualification}, observation 11 is treated unfairly over both observations 1 and 2.
We will show that this can not happen with our method.

\subsection{A fair ranking-based classification algorithm}
In this section, we summarize the previous findings and formalize our idea of a fair ranking-based classification algorithm.
The proposed method can: a) fairly rank a given set of observations, b) propose new labels for the given observations, and c) rank and classify previously unseen observations.
For a), we compute $d''$ for all observations and rank them by distance.
After ranking, we reset the indices such that observation $(1)$ has the smallest distance to the North Star and $(N)$ the largest.
For b) and c), we need to define a capacity threshold $\alpha \in (0,1)$.
This could be, for instance, a given admission rate.
Alternatively, we can set $\alpha$ to the share of positive outcomes within $\mathcal{D}$.
Knowing $\alpha$, we can then determine the cutoff point $\nu := \lceil \alpha N \rceil$, such that the top-$\nu$ observations are assigned the positive outcome $(+)$ and the rest is assigned the negative outcome $(-)$.
To infer a predictor, we compute:
\begin{equation}
    \delta := \frac{\left(d''(\nu,\star)+d''(\nu+1,\star)\right)}{2},
\end{equation}
as the average distance of observations $(\nu)$ and $(\nu +1)$ to the North Star.
Note that $(\nu)$ is the last observation with positive outcome, and $(\nu +1)$ is the first observation with negative outcome.

Ultimately, to classify a previously unseen observation $(u)$, we need to scale its legitimate features according to Algorithm \ref{algo:scaling}---using the $\min$ and $\max$ feature values as observed in $\mathcal{D}$---and measure the distance $d''(u,\star)$ to the North Star.
The inferred predictor would then be:
\begin{equation}
    \widehat{Y}^{(u)} =
    \begin{cases}
        (+) & \textbf{if } d''(u,\star) \leq \delta \\
        (-) & \textbf{otherwise}.
    \end{cases}
\end{equation}
The proposed method is summarized in Algorithm \ref{algo:classifier}.
Note that from Proposition \ref{theorem:more_qualified_shorter_distance}, it follows that meritocratic unfairness can not occur with our method:

\begin{corollary}\label{cor:indiv_unfairness_not}
Meritocratic unfairness, as stated in Definition \ref{def:indiv_unfairness}, can \emph{not} occur if observations are ranked and classified as in Algorithm \ref{algo:classifier}.
\end{corollary}

\begin{proof}
From Proposition \ref{theorem:more_qualified_shorter_distance}, we conclude that if $(i)$ is more qualified than $(j)$, then $d''(i,\star)$ will be strictly smaller than $d''(j,\star)$. But by construction of the ranking in Algorithm \ref{algo:classifier}, we will then have $(i)$ ranked higher than $(j)$, which also implies that if $(j)$ is assigned $(+)$, then $(i)$ as well.
\end{proof}

\begin{algorithm}
\SetKwInOut{Input}{Input}
\SetKwInOut{Output}{Output}
\SetAlgoLined
\Input{Data set $\mathcal{D}$; ($\uparrow$) or ($\downarrow$) relationships for legitimate features; threshold $\alpha$.}
\Output{Ranked and classified observations $(1),\dots,(N)$; predictor $\widehat{Y}$.}
Compute $Z$ as in Algorithm \ref{algo:scaling}\;
Set $z_\ell^{(\star)} \leftarrow 1\quad \forall \ell \in \{1,\dots,L\}$\;
Obtain $\omega_1,\dots,\omega_L$ from learned classifier\;
\For{$\ell \in \{1,\dots,L\}$}{
    $\widetilde{\rho}_\ell \leftarrow \max_{k \in \{1,\dots,K\}} \left\{\left|\rho_{a_k,z_\ell}\right|\right\}$ as in Equation (\ref{eq:max_rho})\;
}
\For{$i \in \{1,\dots,N\}$}{
    $d''(i,\star) \leftarrow \sum_{\ell=1}^L \omega_\ell \left(1-\widetilde{\rho}_\ell\right) \left(1-z_\ell^{(i)}\right)$ as in Equation (\ref{eq:distance_mod_2})\;
    Assign observation $(i)$ the distance $d''(i,\star)$\;
}
Rank observations by distance $d''$ and reset indices such that (1) has smallest distance\;
Define cutoff point $\nu \leftarrow \lceil \alpha N \rceil $\;
Assign $(+)$ to top-$\nu$ observations and $(-)$ to rest\;
Define $\delta \leftarrow \frac{\left(d''(\nu,\star)+d''(\nu+1,\star)\right)}{2}$\;
\eIf{$d''(u,\star) \leq \delta$ for a (potentially unseen) scaled observation $(u)$}{$\widehat{Y}^{(u)} \leftarrow (+)$\;}
{$\widehat{Y}^{(u)} \leftarrow (-)$\;}
\caption{Fair ranking-based classification algorithm.}
\label{algo:classifier}
\end{algorithm}

\subsection{On the relationship to fairness through awareness}
As explained in Section \ref{sec:background}, FTA \cite{dwork2012fairness} is one of the most prominent concepts of individual fairness, which is often verbalized as ``treating similar individuals similarly.''
However, it is often not immediately clear how to measure \emph{similarity} of individuals.
Algorithm \ref{algo:classifier} ranks observations based on their (weighted) distance to the North Star, $d''(\cdot,\star)$.
Hence, by construction, if observations $(i)$ and $(j)$ have (relatively) similar distances to the the North Star, then their rankings will be similar as well.
Specifically, for observations $(i)$, $(j)$, $(k)$, and $rk_{(i)}>rk_{(j)}>rk_{(k)}$, with $rk$ denoting the ranking of an observation, the following two inequalities will always hold:
\begin{align}
    &d''(k,\star) - d''(i,\star) > d''(j,\star) - d''(i,\star) \\
    &d''(k,\star) - d''(i,\star) > d''(k,\star) - d''(j,\star).
\end{align}
However, having a similar distance to the North Star---hence, a similar ranking---does \emph{not} imply that the respective observations are similar in a literal sense.
For instance, in our running graduate school admission example, the two observations in Table \ref{tab:same_distance_to_north_star} would have the same distance to the North Star, despite being fundamentally different in their feature values of GRE V and GRE AW.
\begin{table}[t]
  \caption{Two observations with equal distance to the North Star.}
  \label{tab:same_distance_to_north_star}
  \centering
  \begin{tabular}{l c c c }
    \toprule
    & GRE V & GRE Q & GRE AW \\
    \midrule
    Observation $(i)$ & 170 & 160 & 3.0 \\
    Observation $(j)$ & 140 & 160 & 6.0 \\
    \bottomrule
  \end{tabular}
\end{table}
We argue that this is a desirable property, as it allows individuals with heterogeneous (but equally important/desirable) skill sets to achieve the positive outcome---at least to the extent that $\omega_\ell$ and $\widetilde{\rho}_\ell$, $\ell \in \{1,\dots,L\}$, allow.

On the other hand, it would be difficult to justify an algorithm that assigns significantly different outcomes to similar individuals.
This is, in fact, the reasoning behind FTA.
We will now show that our proposed method respects this requirement---precisely, that similar individuals are guaranteed to have similar distances to the North Star, and thus, similar rankings.
%
But first, we define the similarity of two observations in terms of their weighted distance to each other in the feature space.

\begin{definition}[Similarity of two observations]
We measure the similarity of two observations $(i)$ and $(j)$ by their (weighted) taxicab distance to each other, similar to Equation (\ref{eq:distance_mod_2}):
\begin{equation}
    d''(i,j) := \sum_{\ell=1}^L \omega_\ell \left(1-\widetilde{\rho}_\ell\right)\cdot \left|z_\ell^{(i)}-z_\ell^{(j)}\right|.
\end{equation}
\end{definition}

Again, we have the symmetry $d''(i,j)=d''(j,i)$.
%
%
The following proposition now says that if two observations $(i)$ and $(j)$ are $\varepsilon$-similar, then the difference in their respective distances to the North Star will be bounded by $\varepsilon$.

\begin{proposition}

If two observations $(i)$ and $(j)$ are $\varepsilon$-similar, that is, $d''(i,j) = \varepsilon$, $\varepsilon \geq 0$, then the following holds:
\begin{equation*}
    \left| d''(i,\star) - d''(j,\star)\right| \leq \varepsilon.
\end{equation*}
\end{proposition}


\begin{proof}
%

Let $d''(i,j) = \varepsilon$, and define $\psi_\ell:=\omega_\ell \left(1-\widetilde{\rho}_\ell\right)$. Then we have:
\begin{equation*}
    d''(i,j) = \sum_{\ell=1}^L \psi_{\ell} \cdot \left|z_\ell^{(i)}-z_\ell^{(j)}\right| = \varepsilon.
\end{equation*}
And further, with $\psi_\ell \geq 0$ and the triangle inequality:
\begin{align*}
    \left| d''(i,\star)-d''(j,\star)\right| &= \left|\sum_{\ell=1}^L \psi_\ell \left(1-z_\ell^{(i)}\right) - \sum_{\ell=1}^L \psi_\ell \left(1-z_\ell^{(j)}\right) \right| \\
    &= \left| \sum_{\ell=1}^L \psi_\ell z_\ell^{(i)} - \sum_{\ell=1}^L \psi_\ell z_\ell^{(j)} \right| \\
    &=\left| \sum_{\ell=1}^L \psi_\ell \left(z_\ell^{(i)} - z_\ell^{(j)} \right) \right| \\
    &\leq \sum_{\ell=1}^L \psi_{\ell} \cdot \left|z_\ell^{(i)}-z_\ell^{(j)}\right| = \varepsilon.
\end{align*}
This shows the desired result.
\end{proof}

Now, if we let $\varepsilon$ become small, that is, $\varepsilon \to 0$, then the observations \emph{and} their respective distances to the North Star are becoming increasingly similar---and in the limit equal.
Hence, those observations will be ranked adjacently, everything else unchanged.

\section{Case study: German Credit data set}\label{sec:case_study}
In this section, we instantiate our proposed method on the widely-used German Credit data set \cite{Dua:2019} from 1994.
The data set is made up of 1,000 observations classified as \emph{good} (70\%) or \emph{bad} (30\%) credits ($Y$).
As summarized by \citet{pedreshi2008discrimination}, it includes 20 features on a) personal belongings (e.g., \emph{checking account status}, \emph{savings status}, \emph{property}), b) past/current credits and requested credit (e.g., \emph{credit history}, \emph{credit request amount}), c) employment status (e.g., \emph{job type}, \emph{employment since}), and d) personal attributes (e.g., \emph{personal status and gender}, \emph{age}, \emph{foreign worker}).
%

\begin{table*}[ht]
  \caption{Features of the German Credit data set after pre-processing.}
  \label{tab:german_features}
  \centering
  \begin{tabular}{llcccc}
    \toprule
    Feature & Description & $A$ or $X$ & ($\uparrow$) or ($\downarrow$) & $\omega_\ell$ & $\widetilde{\rho}_\ell$ \\
    \midrule
    \emph{personal status and gender}   & marital status and gender  & $A$ & -- & -- & -- \\
    \emph{age}                          & age of person  & $A$ & -- & -- & --\\
    \emph{foreign worker}               & foreign worker yes/no  & $A$ & -- & -- & --\\
    \midrule
    \emph{checking account status}      & money in checking         & $X$ & ($\uparrow$)    & $0.28$ ($\sigma=0.047$) & $0.07$ \\
    \emph{savings status}               & money in savings          & $X$ & ($\uparrow$)    & $0.11$ ($\sigma=0.029$) & $0.04$ \\
    \emph{property}                     & value of property         & $X$ & ($\uparrow$)    & $0.13$ ($\sigma=0.039$) & $0.13$ \\
    \emph{type of housing}              & free/rent/own             & $X$ & ($\uparrow$)    & $0.06$ ($\sigma=0.025$) & $0.07$ \\
    \emph{credit history}               & quality of credit history & $X$ & ($\uparrow$)    & $0.11$ ($\sigma=0.032$) & $0.15$ \\
    \emph{credit request amount}        & credit amount requested   & $X$ & ($\downarrow$)  & $0.18$ ($\sigma=0.042$) & $0.05$ \\
    \emph{job type}                     & unempl./un-/skilled/mgmt. & $X$ & ($\uparrow$)    & $0.04$ ($\sigma=0.018$) & $0.11$ \\
    \emph{employment since}             & how long employed         & $X$ & ($\uparrow$)    & $0.09$ ($\sigma=0.031$) & $0.32$ \\
    \bottomrule
  \end{tabular}
\end{table*}

From the original data set, we exclude certain features---such as \emph{telephone}---from consideration as they do not exhibit an obvious monotonic relationship with the outcome and, more importantly, appear to be irrelevant for deciding whether to grant a loan or not.
The remaining features are shown in Table \ref{tab:german_features}.
Similar to existing literature, we further separate the remaining features into protected and legitimate features.
We determine the relationships ($\uparrow$ or $\downarrow$) as depicted in Table \ref{tab:german_features}.
For evaluation purposes later on, we randomly shuffle the data and set aside 200 observations for testing purposes, 150 of which are labeled as having \emph{good} credit.

\paragraph{Experimental setup}
First, we scale the legitimate features $X$ as in Algorithm \ref{algo:scaling}.
Then, we fit a random forest classifier with bootstrapping to predict $Y$ from $X$.
We repeat this five times, and for each model, we randomly permute the features ten times---this results in 50 estimates of importance for each legitimate feature.
The average numbers (including standard deviations) are displayed as $\omega_\ell$ in Table \ref{tab:german_features}.
Note that the values of feature importance are only meaningful if the underlying model predicts $Y$ reasonably well.
In our case, we obtain average accuracies of 79.4\% (training) and 78.7\% (testing).
Following Algorithm \ref{algo:classifier}, we next compute the maximum absolute rank correlations $\widetilde{\rho}_\ell$ for each legitimate feature (see Table \ref{tab:german_features}).

We conduct several experiments to rank 200 test observations and predict \emph{good} or \emph{bad} credit.
To that end, we train a 
logistic regression classifier for the following scenarios:
a) using all available features, including protected features (\texttt{LogReg all}), and b) omitting protected features (\texttt{LogReg FTU}).
Third, c) we apply our proposed method to the test observations.

\paragraph{Evaluation criteria and results}
To evaluate the results from scenarios a)--c), we first compare the rankings induced by the respective methods: For a) and b), we rank observations based on the prediction probabilities returned by the classifier, and for c), the ranking is obtained as in Algorithm \ref{algo:classifier}.
We measure fairness of a ranking by the number of unfairly treated observations, as specified in Definitions \ref{def:indiv_unfairness} and \ref{def:qualification}.
In general, an observation can be treated unfairly over \emph{more} than one other observation.

For the baseline models, we therefore measure both the share $\mathcal{S}$ of individual observations that are treated unfairly and the total number $\mathcal{T}$ of instances where meritocratic unfairness occurs.
The results are depicted in Table \ref{tab:indiv_unfairness_baseline}.
Additionally, we also provide numbers on the meritocratic unfairness of the test labels---where we assume that observations with \emph{good} credit are ranked higher than observations with \emph{bad} credit.
We note that \texttt{LogReg all} produces both the highest $\mathcal{S}$ and the highest $\mathcal{T}$, and \texttt{LogReg FTU} performs only marginally better.

\begin{table*}
  \caption{Meritocratic unfairness and accuracy of different scenarios for the German Credit data set.}
  \label{tab:indiv_unfairness_baseline}
  \centering
  \begin{tabular}{lcccc}
    \toprule
    & \texttt{LogReg all} & \texttt{LogReg FTU} & Our Method & Test Labels \\
    \midrule
    $\mathcal{S}$ & $57.5\%$ $\left(\frac{115}{200}\right)$ & $56.0\%$ $\left(\frac{112}{200}\right)$ & $0.0\%$ & $14.5\%$ $\left(\frac{29}{200}\right)$ \\
    $\mathcal{T}$ & $616$ & $596$ & $0$ & $222$ \\
    \midrule
    Accuracy & $78.5\%$ & $76.5\%$ & $56.0\%$ & $100\%$ \\
    \bottomrule
  \end{tabular}
\end{table*}

\begin{figure}
\centering
\input{S_over_alpha}
\caption{Share $\mathcal{S}$ of unfairly treated observations over $\alpha$ for different scenarios. Here, $\mathcal{S}$ is calculated based on the predicted \emph{labels}, not the ranking.} \label{fig:S_over_alpha}  
\end{figure}
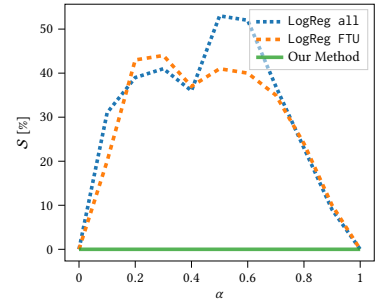

In reality, observations will primarily be affected by the actual outcome of the decision-making task---\emph{good} or \emph{bad} credit.
Hence, we also compare scenarios a)--c) with respect to the predicted outcome, dependent on the choice of a threshold $\alpha$.
Specifically, we calculate $\mathcal{S}$ for $\alpha \in \{0,0.1,0.2,\dots,1\}$ and each scenario.
Note that for the logistic regression models, the positive outcome (\emph{good} credit) is assigned to the $(100\cdot \alpha)\%$ observations with the highest prediction probabilities.
From Figure \ref{fig:S_over_alpha}, we conclude that, apart from the trivial cases of $\alpha=0$ and $\alpha=1$, both baseline models involve high percentages of unfairly treated observations---with \texttt{LogReg all} reaching values of more than 50\% for $\alpha=0.5$ and $\alpha=0.6$.

For completeness, we also include the models' accuracy with respect to the test labels in Table \ref{tab:indiv_unfairness_baseline}.\footnote{We set $\alpha=0.75$ to ensure comparability with the test labels.}
However, note that accuracy is measured based on an imperfect and potentially biased proxy (i.e., the test labels) of the ground-truth labels regarding the qualification of individuals.
Hence, a drop in accuracy, as observed for our method in Table \ref{tab:indiv_unfairness_baseline}, may be explained by a strong mismatch between available imperfect labels and true (but unavailable) labels.
This trade-off between accuracy and fairness is often referred to as the \emph{cost of fairness} \citep{von2021cost,corbett2017algorithmic}.
Unfortunately, we do not have a way to control the level of label bias in real-world data.
For that reason, we conduct a series of experiments on synthetic data and present evidence that, in fact, our method's accuracy tends to be a) similar to traditional classification models' when label bias is low, and b) lower when label bias is high, implying that low accuracy and desirable outcomes need not always contradict each other.

%
%


\section{Experiments on synthetic data}\label{sec:experiments_synthetic}

In order to better understand the previous results, we evaluate our method extensively on synthetic data with imperfect labels.
To that end, we take a more in-depth look at the simplified graduate school admission data introduced in Section \ref{sec:proposed_methodology}.
Recall that we sampled the GRE scores from multivariate Gaussian distributions according to the gender-specific means and standard deviations provided by \citet{ets19,ets18}.
Also, we included an equal amount of females and males, respectively, in the data set of overall 1,000 observations.

\paragraph{Experimental setup}

For the purpose of evaluating our method, we simulate historical admission decisions/labels (e.g., of a technical university) first by computing a score $R$ for each observation as the weighted sum of its (scaled) feature values:
\begin{align*}
    R \coloneqq &\frac{\zeta}{\zeta + 4} \cdot \mathbbm{1}_{male} + \frac{1}{\zeta + 4} \cdot \text{GRE V} + \frac{2}{\zeta + 4} \cdot \text{GRE Q} \\
    &\ + \frac{1}{\zeta + 4} \cdot \text{GRE AW} + \epsilon,
\end{align*}
with $\zeta \geq 0$ and noise $\epsilon \in \mathcal{N}(0,0.1)$, the latter of which might reflect the (unpredictable) mood of the admissions committee or other circumstances that affected admission decisions in the past.
Note that the feature weights sum up to 1, and that the weights of GRE V and GRE AW are the same.
Moreover, the influence of GRE Q on $R$ is approximately twice as high as compared to the other GRE scores, in order to mimic a more quantitative-focused admission process.
The positive outcome $(+)$ is then initially assigned to observations with $R>0.5$, and $(-)$ is assigned otherwise---this ensures a well-balanced label distribution.
%
%
%
Yet, those generated labels are imperfect (i.e., \emph{not} ground truth) because a) the score $R$ is only a noisy signal of potential success in graduate school, b) the computation of $R$ involves (simulated) human subjectivity and error, and c) $R$ may be discriminatory, depending on the choice of $\zeta$.


The parameter $\zeta$ lets us control the amount of \emph{direct} discrimination \cite{mehrabi2019survey} in the decisions, as it directly increases $R$ for males and decreases it for females. 
Besides, a large $\zeta$ could also be an indicator of \emph{indirect} discrimination \cite{mehrabi2019survey}, for instance if other features highly correlated with gender---and favoring males---were given strong weight in the (simulated) historical decisions.
Note that as $\zeta$ becomes increasingly large, the direct influence of the legitimate features (GRE scores) on $R$ vanishes.
%

\begin{table*}
  \caption{Meritocratic unfairness, accuracy, and admission statistics on synthetic data with varying levels of discrimination $\zeta$.}
  \label{tab:indiv_unfairness_synthetic_data_zeta_noise}
  \centering
  \begin{tabular}{llcccc}
    \toprule
    & & \texttt{LogReg all} & \texttt{LogReg FTU} & Our Method & Test Labels \\
    \midrule
    \multirow{6}{*}{$\zeta=0.0$} & $\mathcal{S}$ & $14.0\%$ $\left(\frac{28}{200}\right)$ & $0.0\%$ & $0.0\%$ & $20.0\%$ $\left(\frac{40}{200}\right)$ \\
    & $\mathcal{T}$ & $45$ & $0$ & $0$ & $177$ \\
    \cmidrule(l){2-6}
    & Accuracy & $81.5\%$ & $83.5\%$ & $82.0\%$ ($\alpha=0.59$) & $100.0\%$ \\
    & Admission Female/Male & $0.62$ & $0.72$ & $0.59$ ($\alpha=0.59$) & $0.62$ \\
    & Admission Rate Female & $55.4\%$ & $59.8\%$ & $47.8\%$ ($\alpha=0.59$) & $48.9\%$ \\
    & Admission Rate Male & $75.9\%$ & $70.4\%$ & $68.5$ ($\alpha=0.59$) & $67.6\%$ \\
    \midrule
    \multirow{6}{*}{$\zeta=0.5$} & $\mathcal{S}$ & $42.5\%$ $\left(\frac{85}{200}\right)$ & $0.0\%$ & $0.0\%$ & $18.5\%$ $\left(\frac{37}{200}\right)$ \\
    & $\mathcal{T}$ & $503$ & $0$ & $0$ & $230$ \\
    \cmidrule(l){2-6}
    & Accuracy & $80.5\%$ & $77.5\%$ & $80.0\%$ ($\alpha=0.56$) & $100.0\%$ \\
    & Admission Female/Male & $0.33$ & $0.66$ & $0.53$ ($\alpha=0.56$) & $0.42$ \\
    & Admission Rate Female & $32.6\%$ & $53.3\%$ & $42.4\%$ ($\alpha=0.56$) & $35.9\%$ \\
    & Admission Rate Male & $84.3\%$ & $68.5\%$ & $67.6\%$ ($\alpha=0.56$) & $73.1\%$ \\
    \midrule
    \multirow{6}{*}{$\zeta=1$} & $\mathcal{S}$ & $44.5\%$ $\left(\frac{89}{200}\right)$ & $0.0\%$ & $0.0\%$ & $31.5\%$ $\left(\frac{63}{200}\right)$ \\
    & $\mathcal{T}$ & $1,200$ & $0$ & $0$ & $608$ \\
    \cmidrule(l){2-6}
    & Accuracy & $85.0\%$ & $71.0\%$ & $73\%$ ($\alpha=0.60$) & $100.0\%$ \\
    & Admission Female/Male & $0.13$ & $0.62$ & $0.58$ ($\alpha=0.60$) & $0.26$ \\
    & Admission Rate Female & $14.1\%$ & $48.9\%$ & $47.8\%$ ($\alpha=0.60$) & $27.2\%$ \\
    & Admission Rate Male & $91.7\%$ & $67.6\%$ & $70.4\%$ ($\alpha=0.60$) & $88.0\%$ \\
    \midrule
    \multirow{6}{*}{$\zeta=1.5$} & $\mathcal{S}$ & $44.5\%$ $\left(\frac{89}{200}\right)$ & $18.0\%$ $\left(\frac{36}{200}\right)$ & $0.0\%$ & $36.0\%$ $\left(\frac{72}{200}\right)$ \\
    & $\mathcal{T}$ & $1,571$ & $48$ & $0$ & $797$ \\
    \cmidrule(l){2-6}
    & Accuracy & $90.5\%$ & $68.5\%$ & $71.0\%$ ($\alpha=0.56$) & $100.0\%$ \\
    & Admission Female/Male & $0.03$ & $0.55$ & $0.53$ ($\alpha=0.56$) & $0.14$ \\
    & Admission Rate Female & $3.3\%$ & $41.3\%$ & $42.4\%$ ($\alpha=0.56$) & $15.2\%$ \\
    & Admission Rate Male & $96.3\%$ & $63.9\%$ & $67.6\%$ ($\alpha=0.56$) & $90.7\%$ \\
    \midrule
    \multirow{6}{*}{$\zeta=2$} & $\mathcal{S}$ & $44.5\%$ $\left(\frac{89}{200}\right)$ & $54.5\%$ $\left(\frac{109}{200}\right)$ & $0.0\%$ & $37.5\%$ $\left(\frac{75}{200}\right)$ \\
    & $\mathcal{T}$ & $1,630$ & $324$ & $0$ & $955$ \\
    \cmidrule(l){2-6}
    & Accuracy & $92.5\%$ & $68.0\%$ & $68.0\%$ ($\alpha=0.55$) & $100.0\%$ \\
    & Admission Female/Male & $0.01$ & $0.52$ & $0.54$ ($\alpha=0.55$) & $0.09$ \\
    & Admission Rate Female & $1.1\%$ & $41.3\%$ & $41.3\%$ ($\alpha=0.55$) & $9.8\%$ \\
    & Admission Rate Male & $99.1\%$ & $67.6\%$ & $65.7\%$ ($\alpha=0.55$) & $92.6\%$ \\
    \midrule
    \multirow{6}{*}{$\zeta=2.5$} & $\mathcal{S}$ & $44.5\%$ $\left(\frac{89}{200}\right)$ & $71.0\%$ $\left(\frac{142}{200}\right)$ & $0.0\%$ & $40.5\%$ $\left(\frac{81}{200}\right)$ \\
    & $\mathcal{T}$ & $1,631$ & $731$ & $0$ & $1,221$ \\
    \cmidrule(l){2-6}
    & Accuracy & $95.0\%$ & $68.0\%$ & $65.0\%$ ($\alpha=0.54$) & $100.0\%$ \\
    & Admission Female/Male & $0.00$ & $0.47$ & $0.54$ ($\alpha=0.54$) & $0.05$ \\
    & Admission Rate Female & $0.0\%$ & $37.0\%$ & $41.3\%$ ($\alpha=0.54$) & $5.4\%$ \\
    & Admission Rate Male & $100.0\%$ & $66.7\%$ & $64.8\%$ ($\alpha=0.54$) & $95.4\%$ \\
    \midrule
    \multirow{6}{*}{$\zeta=3$} & $\mathcal{S}$ & $44.5\%$ $\left(\frac{89}{200}\right)$ & $79.5\%$ $\left(\frac{159}{200}\right)$ & $0.0\%$ & $41.5\%$ $\left(\frac{83}{200}\right)$ \\
    & $\mathcal{T}$ & $1,631$ & $1,364$ & $0$ & $1,257$ \\
    \cmidrule(l){2-6}
    & Accuracy & $96.5\%$ & $69.0\%$ & $62.0\%$ ($\alpha=0.54$) & $100.0\%$ \\
    & Admission Female/Male & $0.00$ & $0.42$ & $0.62$ ($\alpha=0.54$) & $0.03$ \\
    & Admission Rate Female & $0.0\%$ & $32.6\%$ & $44.6\%$ ($\alpha=0.54$) & $3.3\%$ \\
    & Admission Rate Male & $100.0\%$ & $65.7\%$ & $61.1\%$ ($\alpha=0.54$) & $96.3\%$ \\
    \bottomrule
  \end{tabular}
\end{table*}

\begin{table*}[ht]
  \caption{Overview of seven synthetic data sets with varying levels of discrimination $\zeta$.}
  \label{tab:synthetic_data_discrimination_omega_rho}
  \centering
  \begin{tabular}{lcc c c c c c c c c}
    & & & $\zeta=0.0$ & $\zeta=0.5$ & $\zeta=1.0$ & $\zeta=1.5$ & $\zeta=2.0$ & $\zeta=2.5$ & $\zeta=3.0$ \\
    \toprule
    Feature & $A$ or $X$ & ($\uparrow$) or ($\downarrow$) & $\omega_\ell$ & $\omega_\ell$ & $\omega_\ell$ & $\omega_\ell$ & $\omega_\ell$ & $\omega_\ell$ & $\omega_\ell$ & $\widetilde{\rho}_\ell$ \\
    \midrule
    \emph{gender} & $A$ & -- & -- & -- & -- & -- & -- & -- & -- & -- \\
    \midrule
    \emph{GRE V} & $X$ & ($\uparrow$) & $0.26$ & $0.21$ & $0.22$ & $0.21$ & $0.22$ & $0.24$ & $0.27$ ($\sigma=0.04$) & $0.04$ \\
    \emph{GRE Q} & $X$ & ($\uparrow$) & $0.66$ & $0.71$ & $0.72$ & $0.70$ & $0.63$ & $0.56$ & $0.49$ ($\sigma=0.05$) & $0.24$ \\
    \emph{GRE AW} & $X$ & ($\uparrow$) & $0.08$ & $0.08$ & $0.07$ & $0.09$ & $0.16$ & $0.20$ & $0.24$ ($\sigma=0.03$) & $0.13$ \\
    \bottomrule
  \end{tabular}
\end{table*}

We evaluate our method on seven synthetic data sets with varying levels of bias/discrimination in the labels, as controlled through $\zeta$ (see Table \ref{tab:indiv_unfairness_synthetic_data_zeta_noise}).
Like in Section \ref{sec:case_study}, we randomly set aside 200 observations for testing on each data set.
Our method is implemented according to Algorithm \ref{algo:classifier}, with the GRE features being legitimate, and gender being protected.
Naturally, higher GRE scores should be more beneficial towards being admitted---hence ($\uparrow$) relationships with the outcome.
An overview of all $\omega_\ell$ and $\widetilde{\rho}_\ell$ values is given in Table \ref{tab:synthetic_data_discrimination_omega_rho}.
Note that $\widetilde{\rho}_\ell$ is constant across the data sets, as changing $\zeta$ only affects the label distribution, not the correlations among features.
We also like to highlight that feature importances ($\omega_\ell$) still capture well the policy that GRE Q should carry significantly more weight in the decision process than GRE V and GRE AW---even with relatively high levels of bias in the labels ($\zeta=3.0$).

\paragraph{Results and interpretation}
As previously, we first compare the rankings of our method against the rankings induced by the logistic regression models \texttt{LogReg all} and \texttt{LogReg FTU} on the test data.
The resulting levels of meritocratic unfairness (both $\mathcal{S}$ and $\mathcal{T}$) are displayed in Table \ref{tab:indiv_unfairness_synthetic_data_zeta_noise}, including the statistics for the test labels.
We also report accuracy as well as additional statistics regarding the admission of females and males for each scenario, based on label predictions.
Specifically, we report the ratio of admitted females to males and compare the admission rates per gender.
For label predictions with our method, to ensure comparability, we set $\alpha$ equal to the share of admitted applicants in the test set.
%

\begin{figure*}[ht]
    \centering
    \begin{minipage}[t]{0.3\textwidth}
        \centering
        \input{S_over_zeta}
        \caption{Share $\mathcal{S}$ of unfairly treated observations over $\zeta$ for different scenarios.}
        \label{fig:S_over_zeta}
    \end{minipage}\hfill
    \begin{minipage}[t]{0.3\textwidth}
        \centering
        \input{Acc_over_zeta}
        \caption{Accuracy over $\zeta$ for different scenarios.}
        \label{fig:Acc_over_zeta}
    \end{minipage}\hfill
    \begin{minipage}[t]{0.3\textwidth}
        \centering
        \input{AR_over_zeta}
        \caption{Admission ratio of females to males over $\zeta$ for different scenarios.}
        \label{fig:AR_over_zeta}
    \end{minipage}
\end{figure*}

From Table \ref{tab:indiv_unfairness_synthetic_data_zeta_noise} and Figures \ref{fig:S_over_zeta}--\ref{fig:AR_over_zeta}, we infer several observations: First, as $\zeta$ increases, the admission ratio of females to males in the data sets decreases, as expected, whereas the overall admission rate remains stable (between $54\%$ and $60\%$ in the test labels).
The fact that increasing $\zeta$ results in the labels depending stronger on the value of \emph{gender} is exploited by \texttt{LogReg all} to discriminate observations based thereon.
Not surprisingly, as $\zeta$ increases, \texttt{LogReg all} clings to the trajectory of the test labels both for accuracy as well as meritocratic unfairness and demographic parity (with respect to admission ratios), making its predictions accurate but blatantly unfair---both meritocratically and with respect to admission rates by gender.
We further observe in Figure \ref{fig:S_over_zeta} that the gender-agnostic \texttt{LogReg FTU} model removes meritocratic unfairness when label bias is low ($\zeta<1.0$).
However, as the level of discrimination in the data increases, it fails to remove such unfairness, with $\mathcal{S}$ surging and even surpassing the levels of \texttt{LogReg all} and the test labels for growing $\zeta$.
These problems do not occur with our method, which always achieves zero meritocratic unfairness.
Additionally, as can be seen in Figure \ref{fig:AR_over_zeta}, our experiments suggest that enforcing individual meritocratic fairness results in higher group fairness (here: demographic parity with respect to admission rates) as well: While the logistic regression models both exhibit a negative relationship between $\zeta$ and demographic parity, our method satisfies a constant high level of group fairness for any $\zeta$, similar to the one of the test labels without explicit discrimination ($0.62$ for $\zeta=0.0$).
Note that the converse---group fairness implying individual meritocratic fairness---is not generally true, for instance if a model randomly admits an equal share of females and males without paying any attention to their qualification.
\balance

\section{Conclusion}
\label{sec:conclusion}
In this paper, we present a practical and easy-to-implement approach for fair ranking and binary classification
based on monotonic relationships between legitimate features (or interactions thereof) and the outcome.
Given the common setting of data with (potentially biased) imperfect labels, our method ranks observations according to their qualification for a specific outcome, for instance admission to graduate school, regardless of protected features like gender or race.
Instead of learning to predict imperfect labels, we introduce an idea to incorporate useful information from historical decisions in our decision criterion.
Additionally, we account for unwanted dependencies between (seemingly) legitimate and protected features.
We show theoretically that our method respects a version of the prominent concept of \emph{fairness through awareness}.
Experiments on synthetic and real-world data confirm that our method yields desirable results both with respect to meritocratic fairness and group fairness (e.g., similar admission rates for females and males), clearly outperforming traditional classification algorithms trained on data with biased/imperfect labels.

Our work involves certain limitations that allow for various directions of follow-up research: For instance, it would be interesting to elaborate more on how to meaningfully include features that do not exhibit obvious monotonic relationships with the outcome.
Another natural extension of our method could involve a more sophisticated and natural way of accounting for feature interactions.
Perhaps most importantly, we stress that our proposed method satisfies the introduced conception of meritocratic fairness as well as demographic parity (empirically), but may \emph{not} be fair according to other notions of fairness.
More generally, we like to highlight that fairness in the societal sense cannot be reduced to a simple technical property.
For instance, understanding people's fairness \emph{perceptions} towards algorithmic decision systems is another vital research endeavor \cite{lee2018understanding,schoffer2021study}.
Hence, it is necessary to have different notions of fairness---and our work may be seen as one contribution to this toolbox.
%
Ultimately, we hope that our work will equip (especially) practitioners with helpful new tools for designing equitable decision systems.

\begin{acks}
We like to thank our anonymous reviewers for their thoughtful and constructive feedback, which helped strengthen this manuscript.
\end{acks}

\bibliographystyle{ACM-Reference-Format}
\bibliography{bibliography}

\appendix


\end{document}

%% file: S_over_alpha.tex



\begin{tikzpicture}[scale=0.6]

\definecolor{color0}{rgb}{0.12156862745098,0.466666666666667,0.705882352941177}
\definecolor{color1}{rgb}{1,0.498039215686275,0.0549019607843137}
\definecolor{color2}{rgb}{0.172549019607843,0.627450980392157,0.172549019607843}

\begin{axis}[
legend cell align={left},
legend style={fill opacity=0.5, draw opacity=1, text opacity=1, draw=white!80!black},
tick align=outside,
tick pos=left,
x grid style={white!69.0196078431373!black},
xlabel={$\alpha$},
xmin=-0.05, xmax=1.05,
xtick style={color=black},
y grid style={white!69.0196078431373!black},
ylabel={$\mathcal{S}$ [\%]},
ymin=-2.65, ymax=55.65,
ytick style={color=black}
]
\addplot [line width = 0.8mm, dotted, color0]
table {%
0 0
0.1 31
0.2 39
0.3 41
0.4 36
0.5 53
0.6 52
0.7 37
0.8 23
0.9 9
1 0
};
\addlegendentry{\texttt{LogReg all}}
\addplot [line width = 0.8mm, dashed, color1]
table {%
0 0
0.1 20
0.2 43
0.3 44
0.4 37
0.5 41
0.6 40
0.7 35
0.8 24
0.9 10
1 0
};
\addlegendentry{\texttt{LogReg FTU}}
\addplot [line width = 0.8mm, color2, draw opacity = 0.8]
table {%
0 0
0.1 0
0.2 0
0.3 0
0.4 0
0.5 0
0.6 0
0.7 0
0.8 0
0.9 0
1 0
};
\addlegendentry{Our Method}
\end{axis}

\end{tikzpicture}


%% file: S_over_zeta.tex
\begin{tikzpicture}[scale=0.6]

\definecolor{color0}{rgb}{0.12156862745098,0.466666666666667,0.705882352941177}
\definecolor{color1}{rgb}{1,0.498039215686275,0.0549019607843137}
\definecolor{color2}{rgb}{0.172549019607843,0.627450980392157,0.172549019607843}

\begin{axis}[
legend cell align={left},
legend pos=north west,
legend style={fill opacity=0.5, draw opacity=1, text opacity=1, draw=white!80!black},
tick align=outside,
tick pos=left,
x grid style={white!69.0196078431373!black},
xlabel={$\zeta$},
xmin=-0.05, xmax=3.05,
xtick style={color=black},
y grid style={white!69.0196078431373!black},
ylabel={$\mathcal{S}$ [\%]},
ymin=-2.65, ymax=82.65,
ytick style={color=black}
]
\addplot [line width = 0.8mm, dotted, color0]
table {%
0   14
0.5 42.5
1   44.5
1.5 44.5
2   44.5
2.5 44.5
3   44.5
};
\addlegendentry{\texttt{LogReg all}}
\addplot [line width = 0.8mm, dashed, color1]
table {%
0   0
0.5 0
1   0
1.5 18
2   54.5
2.5 71
3   79.5
};
\addlegendentry{\texttt{LogReg FTU}}
\addplot [line width = 0.8mm, color2, draw opacity = 0.8]
table {%
0   0
0.5 0
1   0
1.5 0
2   0
2.5 0
3   0
};
\addlegendentry{Our Method}
\addplot [line width = 0.8mm, loosely dotted, black]
table {%
0   20
0.5 18.5
1   31.5
1.5 36
2   37.5
2.5 40.5
3   41.5
};
\addlegendentry{Test Labels}
\end{axis}

\end{tikzpicture}

%% file: Acc_over_zeta.tex
\begin{tikzpicture}[scale=0.6]

\definecolor{color0}{rgb}{0.12156862745098,0.466666666666667,0.705882352941177}
\definecolor{color1}{rgb}{1,0.498039215686275,0.0549019607843137}
\definecolor{color2}{rgb}{0.172549019607843,0.627450980392157,0.172549019607843}

\begin{axis}[
legend cell align={left},
legend pos=south west,
legend style={fill opacity=0.5, draw opacity=1, text opacity=1, draw=white!80!black},
tick align=outside,
tick pos=left,
x grid style={white!69.0196078431373!black},
xlabel={$\zeta$},
xmin=-0.05, xmax=3.05,
xtick style={color=black},
y grid style={white!69.0196078431373!black},
ylabel={Accuracy [\%]},
ymin=47.35, ymax=102.65,
ytick style={color=black}
]
\addplot [line width = 0.8mm, dotted, color0]
table {%
0   81.5
0.5 80.5
1   85
1.5 90.5
2   92.5
2.5 95
3   96.5
};
\addlegendentry{\texttt{LogReg all}}
\addplot [line width = 0.8mm, dashed, color1]
table {%
0   83.5
0.5 77.5
1   71
1.5 68.5
2   68
2.5 68
3   69
};
\addlegendentry{\texttt{LogReg FTU}}
\addplot [line width = 0.8mm, color2, draw opacity = 0.8]
table {%
0   82
0.5 80
1   73
1.5 71
2   68
2.5 65
3   62
};
\addlegendentry{Our Method}
\addplot [line width = 0.8mm, loosely dotted, black]
table {%
0   100
0.5 100
1   100
1.5 100
2   100
2.5 100
3   100
};
\addlegendentry{Test Labels}
\end{axis}

\end{tikzpicture}

%% file: AR_over_zeta.tex
\begin{tikzpicture}[scale=0.6]

\definecolor{color0}{rgb}{0.12156862745098,0.466666666666667,0.705882352941177}
\definecolor{color1}{rgb}{1,0.498039215686275,0.0549019607843137}
\definecolor{color2}{rgb}{0.172549019607843,0.627450980392157,0.172549019607843}

\begin{axis}[
legend cell align={left},
legend pos=south west,
legend style={fill opacity=0.5, draw opacity=1, text opacity=1, draw=white!80!black},
tick align=outside,
tick pos=left,
x grid style={white!69.0196078431373!black},
xlabel={$\zeta$},
xmin=-0.05, xmax=3.05,
xtick style={color=black},
y grid style={white!69.0196078431373!black},
ylabel={Admission Female/Male},
ymin=-0.05, ymax=0.85,
ytick style={color=black}
]
\addplot [line width = 0.8mm, dotted, color0]
table {%
0   0.62
0.5 0.33
1   0.13
1.5 0.03
2   0.01
2.5 0
3   0
};
\addlegendentry{\texttt{LogReg all}}
\addplot [line width = 0.8mm, dashed, color1]
table {%
0   0.72
0.5 0.66
1   0.62
1.5 0.55
2   0.52
2.5 0.47
3   0.42
};
\addlegendentry{\texttt{LogReg FTU}}
\addplot [line width = 0.8mm, color2, draw opacity = 0.8]
table {%
0   0.59
0.5 0.53
1   0.58
1.5 0.53
2   0.54
2.5 0.54
3   0.62
};
\addlegendentry{Our Method}
\addplot [line width = 0.8mm, loosely dotted, black]
table {%
0   0.62
0.5 0.42
1   0.26
1.5 0.14
2   0.09
2.5 0.05
3   0.03
};
\addlegendentry{Test Labels}
\end{axis}

\end{tikzpicture}